%% file: main.tex
\documentclass[runningheads]{llncs}

\usepackage[T1]{fontenc}
\usepackage{graphicx}
\usepackage{todonotes}
\usepackage{booktabs}
\usepackage[table]{xcolor}
\usepackage{supertabular}
\usepackage{adjustbox}
\usepackage{float}
\usepackage{tabularx}
\usepackage{rotating}
\usepackage{tablefootnote}
\usepackage{verbatim}
\usepackage{listings}
\usepackage{hyperref}
\usepackage{lscape}
 \usepackage{amsmath}
 \usepackage{tabularx}
\usepackage{amssymb}
\usepackage{paralist}
\usepackage{wrapfig}

\usepackage{longtable}
\usepackage{microtype}
\usepackage{xltabular}

\usepackage[inline]{enumitem}

\renewcommand{\paragraph}[1]{\textbf{#1}\ }
\begin{document}
%

\title{Characterising LLM-Generated Competency Questions: a Cross-Domain Empirical Study \\ using Open and Closed Models}

\titlerunning{Characterising LLM-Generated Competency Questions}
%
%
%
\author{
Reham Alharbi\inst{1}\orcidID{0000-0002-8332-3803} \and
Valentina Tamma\inst{2}\orcidID{0000-0002-1320-610X} \and
Terry R. Payne\inst{2}\orcidID{0000-0002-0106-8731} \and
Jacopo de Berardinis\inst{2}\orcidID{0000-0001-6770-1969}
}

\authorrunning{R. Alharbi et al.}

\institute{
Taibah University, Madinah, Saudi Arabia \\
 \email{rfalharbi@taibahu.edu.sa}
\and
University of Liverpool, Liverpool, UK \\
\email{\{v.tamma,t.r.payne,jacodb\}@liverpool.ac.uk} 
}

\maketitle              
\begin{abstract}
Competency Questions (CQs) are a cornerstone of requirement elicitation in ontology engineering.
CQs represent requirements as a set of natural language questions that an ontology should satisfy; they are traditionally modelled by ontology engineers together with domain experts as part of a human-centred, manual elicitation process.
The use of Generative AI automates CQ creation at scale, therefore democratising the process of generation, widening stakeholder engagement, and ultimately broadening access to ontology engineering. 
However, given the large and heterogeneous landscape of LLMs, varying in dimensions such as parameter scale, task and domain specialisation, and accessibility, it is crucial to characterise and understand the intrinsic, observable properties of the CQs they produce (e.g., readability, structural complexity) through a systematic, cross-domain analysis.
This paper introduces a set of quantitative measures for the systematic comparison of CQs across multiple dimensions. Using CQs generated from well defined use cases and scenarios, we identify their salient properties, including \emph{readability}, \emph{relevance with respect to the input text} and \emph{structural complexity of the generated questions}.
We conduct our experiments over a set of use cases and requirements using a range of LLMs, including both open (KimiK2-1T, LLama3.1-8B, LLama3.2-3B) and closed models (Gemini 2.5 Pro, GPT 4.1).  
Our analysis demonstrates that LLM performance reflects distinct generation profiles shaped by the use case. 

\keywords{Knowledge engineering  \and Competency questions  \and LLMs}
\end{abstract}

\input{sections/introduction}
\input{sections/related}

\input{sections/methodology}
\input{sections/results}
\input{sections/discussion+conclusion}
%
%
%
\bibliographystyle{splncs04}
\bibliography{references.bib}

\end{document}

%% file: sections/introduction.tex
\section{Introduction}\label{sec:intro}

Requirement elicitation constitutes a foundational stage in the ontology engineering (OE) lifecycle, determining the functional scope and semantic adequacy of the resulting model.
Competency questions (CQs)~\cite{gruninger1995methodology} are recognised as the standard mechanism for this task, serving as a natural language interface between domain experts and ontology engineers.
By framing requirements as answerable questions, CQs direct the modelling of concepts and relations~\cite{SurezFigueroa2015NEON,Presutti2009eXtremeDW}, underpin validation and testing~\cite{Keettestdriven2016,bezerra2017verifying}, and inform assessments of ontology reuse~\cite{Aziz2023,FernandezLopez2019}.
However, the manual formulation of CQs remains a major bottleneck, as it is labour-intensive and requires substantial domain and modelling expertise, which in practice leads to their under-utilisation~\cite{alharbi2021,monfardini2023}.
To mitigate this, the OE community has increasingly turned towards automation, from pattern-based approaches~\cite{CLARO2019,Ren2014,wisniewski2019analysis} to the use of Large Language Models (LLMs)~\cite{alharbi2024SAC,AgoCQ_Keet,RevOnt_2024,rebboud2024_ESWC,Bohui2025}.

This rapid adoption of LLMs in OE has failed to 
understand their output properties across multiple dimensions and characteristics. The landscape of available models is vast, encompassing 
both proprietary ``closed'' and 
``open-weight'' models spanning a range of parameter scales.
While recent studies have demonstrated the feasibility of using LLMs for CQ generation \cite{alharbi2024SAC,RevOnt_2024}, the systematic comparison and evaluation of the intrinsic properties of their outputs  remains a significant challenge; for example, the influence of model architecture, parameter size, and input domain on features such as linguistic structure, complexity, and semantic diversity of generated CQs remains poorly investigated. Treating LLMs as a monolithic solution 
ignores the substantial variability in their output profiles that 
ontology engineers must understand to effectively select LLM based OE tools.

We address this gap by shifting the focus from generation feasibility to a systematic \textit{characterisation} of LLM-generated CQs.
We present a broad, cross-domain empirical study that compares five distinct LLMs -- comprising both state-of-the-art closed models (\texttt{Gemini}, \texttt{GPT}) and open models of varying capacity (\texttt{Llama 3.1-2}, \texttt{KimiK2}) -- across five diverse domains ranging from cultural heritage to healthcare.
We analyse how different models interpret user stories and use cases to produce functional requirements as CQs.
To support this analysis, we introduce \textsc{CompCQ}, a multi-dimensional framework to profile CQs features.

This paper makes a twofold contribution. First, we introduce \textsc{CompCQ}, a comprehensive, multi-dimensional framework designed to systematically characterise and quantify key linguistic, syntactic, and semantic properties of competency questions.
Second, we present a large-scale empirical study that applies \textsc{CompCQ} and analyses and compares CQs generated by diverse LLMs from the same defined user stories and use cases across multiple domains.
The empirical study is performed on a curated, multi-domain dataset of user stories and use cases.

Overall, our results show that domain characteristic are the primary determinant of LLM generation behaviour, driving divergence in broad domains, convergence in narrow ones, and uniformly complex outputs in technically demanding settings. The framework shows how each model exhibits a distinct and consistent generation profile, with closed models offering greater stability and readability and open models trading clarity for higher diversity, while some models fail to provide adequate coverage altogether. Crucially, our result also demonstrate that no single LLM can capture the full requirements space; effective CQ generation therefore requires combining multiple models and retaining human-in-the-loop refinement to ensure comprehensive and accurate coverage.

%% file: sections/related.tex
\section{Related work}\label{sec:related}
 
Competency Questions (CQs)~\cite{uschold1995towards} are used in the ontology development process to scope the ontology, validate its content, and guide the reuse of existing ontological resources~\cite{Alharbi2021AssessingCO,Aziz2023,bezerra2017verifying,Dennis2017,Keettestdriven2016,Kim2007}.
However, CQs are formulated in diverse ways, without a common standard or clear guidelines defining how they should be written, what constitutes a good CQ, or how they should be assessed~\cite{alharbi2024characteristics,alharbi-et-al2024:ekaw,alharbi2021,AgoCQ_Keet,monfardini2023,keet-kahn2024:ekaw}.
The only explicit guidance derived from Gruninger and Fox's definition~\cite{GruningerFox94} is that CQs should specify the requirements for an ontology and, in doing so, characterise its scope.
CQs therefore act as constraints on what the ontology can represent, rather than determining a specific design or set of ontological commitments.

The introduction of filler-based approaches, which incorporate patterns~\cite{Ren2014,wisniewski2019analysis}, templates~\cite{bezerra2017verifying,bezerra2014cqchecker,CLARO2019}, and controlled natural languages~\cite{antia-keet-2021-assessing}, represents an important step toward standardizing CQ formulation.
However, several limitations persist, including the incomplete coverage of filler-based patterns~\cite{antia-keet-2021-assessing,wisniewski2019analysis}, the manual effort required to verify a CQ’s compliance, and the need for manual revision to ensure alignment.
These factors make the process time-consuming, labour-intensive, and prone to grammatical errors~\cite{AgoCQ_Keet}, thereby hindering their broader applicability.

More recently, various efforts have used LLMs to generate domain-specific CQs~\cite{alharbi2024SAC,AgoCQ_Keet,RevOnt_2024,Bohui2025,rebboud2024_ESWC} varying prompts and contextual information. However, there is a lack of consensus on the approaches to use to evaluate the generated CQs. 
A prevalent approach is based on the involvement of human evaluators, typically domain experts, and in comparing the LLM generated CQs against those that are manually curated ~\cite{alharbi2024SAC,rebboud2024_ESWC}.
However, this approach can be problematic due to the subjectivity and variability of manual questions, scalability limitations, challenges in capturing semantic equivalence, domain-specific constraints, and the difficulty in adequately assessing the diversity and novelty of generated outputs.


There is a critical need for systematic methods that enable fair, unbiased, and scalable evaluation frameworks, facilitating robust comparative assessments even when absolute quality judgments remain elusive.
Our work provides a foundation for understanding the characteristic features of LLM-authored CQs by quantifying linguistic, syntactic, and semantic properties of competency questions, and therefore providing a multifaceted characterization essential for a robust evaluation. Linguistic features such as \emph{readability} directly impact the interpretability and usability of CQs by domain experts and ontology engineers, ensuring that generated questions can effectively guide ontology development. Syntactic complexity measures, including \emph{structural depth} and \emph{part-of-speech distributions}, reveal the cognitive and computational demands made by CQs, influencing both human comprehension and automated processing. Meanwhile, semantic properties, assessed through \emph{embedding-based cluster analysis} and \emph{overage metrics}, capture the conceptual relevance and diversity of questions relative to the source domain, which are critical for comprehensive ontology validation and reuse. These dimensions collectively define the systematic framework that enables a fine-grained comparative evaluation of LLM-generated CQs, \texttt{CompCQ}.

%% file: sections/methodology.tex
\section{CompCQ: a framework for comparing CQ features}\label{sec:methodology}


The aim of the \textsc{CompCQ} framework is to quantify the linguistic, structural, and semantic characteristics of LLM-generated questions to support comparative studies, therefore addressing two primary research questions:

\begin{description}[leftmargin=31pt,itemsep=2pt,topsep=2pt]
    \item[\textbf{RQ1:}] What linguistic and structural features distinguish CQs generated by both open and closed LLMs across various domains? We analyse quantitative metrics such as \emph{readability}, \emph{relevance}, and \emph{complexity} to construct a comprehensive data-driven profile of generated questions.
    \item[\textbf{RQ2:}] How do LLM-generated CQ sets compare in terms of internal semantic diversity and inter-model overlap? We utilize embedding-based methods to assess conceptual breadth within sets and measure coverage across different models to identify shared and unique requirements.
\end{description}

\noindent
Our methodology operationalises these questions by focusing on three core CQ-level features, \textit{readability}, \textit{complexity}, and \textit{relevance} to source requirements, as well as set-level metrics that capture \textit{diversity} and \textit{overlap}.
%
This selection bridges established evaluation criteria from natural language processing and ontology engineering. For example, dimensions such as clarity, typically addressed through readability and syntactic complexity, reflect the ease with which ontology engineers interpret and formalise questions. Relevance is assessed relative to input requirements, while completeness and semantic richness emerge through diversity and coverage metrics at the set level.
We contend that the practical usability of CQs can be understood as an emergent property of these intrinsic features.


\subsection{Requirement specifications and case studies}\label{sssec:requirement-dataset}

To evaluate model behaviour across different input structures, we selected requirement specifications in two formats widely adopted in OE: \textit{user stories} and \textit{use cases}.
While both formats serve to capture requirements, they differ fundamentally in structure and focus. \textbf{User stories} are narrative-driven, focusing on the ``who'' (persona) and the ``why'' (goal/benefit), often leaving the specific system interactions implicit \cite{de2023polifonia}.
In contrast, \textbf{use cases} provide a structured, step-by-step description of system behaviour, explicitly detailing the ``how'' through actors, preconditions, and flow of events \cite{kendall2019ontology}. 
We selected five distinct domains -- from cultural heritage to healthcare -- to ensure our analysis captures potential domain-dependent variations in CQ generation.

For user stories, we utilise two scenarios rooted in cultural heritage and metadata, specified via personas and narrative goals:
\begin{inparaenum}[(i)]
    \item\textbf{British Music Experience (BME)} \cite{alharbi2025comparative}: a cultural heritage scenario involving a museum curator and a donor focused on collection management; and
    \item \textbf{Music Meta Ontology} \cite{de2023music}: a scenario featuring a computer scientist, aiming to integrate heterogeneous music datasets for information completion.
\end{inparaenum}
The three use cases selected were transcribed from the \textit{Ontology Engineering} textbook repositories \cite{kendall2019ontology}, representing technical, social, and geospatial domains:
\begin{inparaenum}[(i)]
    \item \textbf{Personalized Depression Treatment Ontology (PDTO)}: a healthcare use case correlating patient demographics with genetic and clinical trial data;   
    \item \textbf{Political Journalism Ontology (PJO)}: a media analysis scenario aimed at detecting reporting trends and bias across different news outlets; and
    \item \textbf{When To Go Where (WTGW)}: a tourism recommender system that suggests national parks based on user preferences. 
\end{inparaenum}

\subsection{LLMs for CQ generation}\label{sssec:llm-cq}

Given the proliferation of LLMs, each exhibiting distinct architectural and functional profiles, we sought to evaluate our feature set across a representative spectrum of model types. To this end, we select a diverse set of LLMs with varying architectures, parameter sizes, and access modalities:

\begin{description}[leftmargin=0pt,itemsep=5pt,topsep=2pt]
\item[\textbf{Closed models:}]
\texttt{Gemini 2.5 Pro} was selected as the current state-of-the-art model in its family (Gemini 3 was only in preview at the time of writing). Likewise, \texttt{GPT-4.1} was selected over GPT-5 to ensure experimental reproducibility, particularly regarding the ability to control hyper-parameters such as temperature.
\item[\textbf{Open models:}] we utilised models via the Ollama library to test varying levels of capacity. These ranged from lightweight models suitable for local inference, such as \texttt{Llama 3.1 8B} and \texttt{Llama 3.2 3B} (Meta) \cite{dubey2024llama}, to large-scale models like \texttt{Kimi-K2} (Moonshot AI). The latter, with 1 trillion parameters, represents the high-capacity end of open models and was accessed via the Ollama cloud API.
\end{description}

To generate CQs, we adopted a neutral (0-shot) \emph{prompting strategy}, providing the LLMs only with the requirement specification and a minimal instruction to \emph{``generate a set of competency questions''} from the text.
We explicitly avoided providing examples or desiderata to minimise any bias.
It is also important to mitigate the issue of \emph{data leakage} (i.e. whether or not the models were exposed to ``gold standard'' CQs during pre-training) through three domain-specific factors.
First, the three use case datasets (PDTO, PJO, WTGW) are derived from academic coursework with no published CQs.
Second, the \textit{BME} dataset is a recently curated resource whose publication date falls beyond the knowledge cutoff of the selected models.
Third, for \textit{Music Meta} (released in Q4-2023), we conducted preliminary adversarial testing and found that even with specific prompting, the models failed to reproduce the exact phrasing or structure of the original human-authored CQs, suggesting that these specific requirements were not part of their training corpus.

\subsection{CQ feature extraction}\label{sssec:cq-features}

For each CQ we extract 3 types of features: \emph{readability}, \emph{complexity}, and \emph{relevance to the requirements}.
Their definition, rationale, and extraction are outlined below.

\input{tables/readability_measures}

\vspace{0.05in} 
\noindent \paragraph{Readability}
is assessed for each CQ dataset
to gauge its ease of understanding.
Since there is no universal definition of readability, and following the approach in~\cite{RevOnt_2024}, we compute a suite of established readability indices, designed to capture different aspects of textual difficulty.
In this paper, we selected the Flesch-Kincaid Grade Level (FKGL) and the Dale-Chall Readability Score (DCR) as representative readability features.
These metrics estimate the US grade level required for comprehension (FKGL) and the reliance on common vocabulary (DCR), as explained in Table~\ref{tab:readabilityMeasures}.
Generally, lower scores indicate better readability, suggesting that less specialist knowledge or lower formal educational attainment is required for comprehension.
It is worth noting that most readability formulae, including FKGL and DCR, were primarily developed for continuous prose rather than short, interrogative sentences like CQs.
Consequently, we interpret the resulting scores as \textit{comparative indicators} of readability across rather than absolute measures.

\vspace{0.05in} 
\noindent \paragraph{Complexity} of a CQ dataset can vary significantly, impacting modelling effort, potential ambiguity, and the intricacy of subsequent data querying.
To gain a quantitative understanding of these variations, we employ a multi-faceted approach that characterises each CQ along three distinct dimensions ($c_1,c_2,c_3$).

\begin{itemize}[leftmargin=0pt,itemsep=2pt,topsep=0pt]
    \renewcommand\labelitemi{}
    \item \textbf{\textit{Requirement complexity} ($c_1$)} quantifies the complexity inherent in the emergence of \textit{ontological primitives} potentially required to represent and retrieve the necessary information.
    This is done by identifying the distinct \texttt{Concepts} (classes/entity types), \texttt{Properties} (data attributes), \texttt{Relationships} (links between concepts), and specific \texttt{Filters} (constraints on properties or class memberships). Additionally, the expected \texttt{Cardinality} (implying retrieval of a single instance, multiple instances, or an existence check) and any \texttt{Aggregation} requirements (such as counting or averaging) are identified. These primitives are expected to reflect the anticipated richness and interconnectedness of the induced ontological model and the likely complexity of corresponding queries. The primitives are extracted by an LLM (\texttt{Gemini 2.5 Pro}) and aggregated through a scoring function that sums the number of occurrences for each metric.

    \item \textbf{\textit{Linguistic complexity} ($c_2$)} quantifies aspects of the question's wording, length, and basic grammatical composition using NLP methods for part-of-speech tagging and noun chunking.
    Metrics are extracted automatically via \texttt{spaCy}, and include counts of \texttt{Noun Phrases} (proxy for entity/argument density), \texttt{Verbs} (actions/states), \texttt{Prepositions} (often indicating relational phrases), \texttt{Coordinating Conjunctions} (suggesting logical combinations), and \texttt{Modifiers} (adjectives/adverbs, reflecting descriptive detail or potential filters). The primary \texttt{Interrogative Structure} (e.g., WH-retrieval, Boolean check, aggregation via ``How many'') is also identified. This analysis provides a measure of the requirement statement's textual elaboration.
    Similarly, a scoring function summing over the metrics count with equal weighting is used to obtain a scalar.

    \item \textbf{\textit{Syntactic complexity} ($c_2$)} focuses on \textit{grammatical structure and relational syntax} by traversing the generated dependency tree for each CQ.
    These include the total \texttt{Node Count} (token length), the maximum dependency path length or \texttt{Tree Depth} (indicating syntactic nesting and long-distance dependencies), and the frequency of specific \texttt{Dependency Relation} types indicative of complex grammatical constructions i.e., \texttt{nsubj}, \texttt{dobj}, \texttt{prep}, \texttt{acl}, \texttt{relcl}, \texttt{conj}, \texttt{agent}, selected based on linguistic complexity heuristics from the Universal Dependency set \cite{de-marneffe-etal-2014-universal}. All counts and depths are aggregated by sum.
\end{itemize}

We also compute the number of characters of each CQ ($c_0$ complexity) as an indicator of verbosity and potential elaboration.
Overall, we expect these four dimensions -- length, requirement, linguistic, and syntactic structure –- to provide complementary perspectives on CQ complexity.
A CQ might be semantically complex (e.g., requiring navigation of intricate partonomy or causality relations) yet linguistically simple (e.g., ``What caused this event?''), scoring high on requirement metrics but low on linguistic/syntactic ones.
Conversely, a CQ might be ontologically straightforward but phrased using complex sentence structures, scoring high on syntactic metrics but low on semantic ones.

\vspace{0.05in} 
\noindent \paragraph{Relevance} measures the degree to which a CQ aligns with the requirements expressed in the user story.
This assessment was performed by an LLM (\texttt{Gemini 2.5 Pro}), which was instructed to rate the relevance of each CQ on a 4-point Likert scale.
The scale was defined to capture different aspects of relevance from an ontology engineering perspective, where higher scores indicate closer alignment with explicitly stated or functionally necessary requirements.
Specifically, the scale points represented the following:
(4) \textit{The CQ addresses a requirement explicitly stated in the text};
(3) \textit{The CQ pertains to a requirement that, while not explicit, is inferable from the requirements using domain knowledge and is functionally necessary for fulfilling the story's goals};
(2) \textit{The CQ is not directly inferable from the requirements but holds some contextual relevance to the persona or their objectives within the requirements};
(1) \textit{The CQ introduces an additional requirement that is neither expressed in, nor inferable from, the requirements and is not considered necessary}.
A subset of these LLM judgments (12 CQs, with their expected ratings) were manually validated for prompt engineering.


\subsection{Semantic analysis of CQ sets via sentence embeddings}
To analyse the semantic characteristics of CQ sets generated by different methodologies, we conducted two main studies.
These studies utilised Sentence-BERT embeddings from the \texttt{all-MiniLM-L6-v2} model, which generates vectors $\mathbf{e} \in \mathbb{R}^{384}$ capturing the semantic meaning of each CQ, as done in \cite{alharbi-et-al2024:ekaw,RevOnt_2024}.

\vspace{0.05in} 
\noindent \paragraph{Internal semantic diversity}
quantifies the internal diversity of each CQ set as intra-cluster variability \cite{tevet2021evaluating}.
This is achieved through the following metrics.

\begin{itemize}[leftmargin=0pt,itemsep=2pt,topsep=0pt]
    \renewcommand\labelitemi{}
    \item \textbf{\textit{Average pairwise cosine similarity (APS)}} between all unique pairs of CQ embeddings $(\mathbf{e}_i, \mathbf{e}_j)$ in the set. A \textit{lower mean} suggests \textit{higher diversity} (CQs are, on average, less semantically similar, implying a broader range of topics).

    \item \textbf{\textit{Average centroid distance (ACD)}} measures the Euclidean distance of all embeddings to the set centroid (mean vector $\mathbf{\bar{e}}$). Intuitively, this measures the semantic dispersion of CQs around the set's central theme; a \textit{higher mean} indicates \textit{greater diversity} (CQs are more spread out in the embedding space).

    \item \textbf{\textit{Shannon entropy ($H$)}} estimates the breadth of topics covered. We first clustered the set embeddings via k-means clustering, partitioning them into $k$ semantic clusters ($k=5$ was chosen given that each set comprises 8-35 CQs from one user story/case).
    The Shannon entropy was then computed on the probability distribution ($p_c$) of CQs across these $k$ clusters using the formula $H = -\sum_{c=1}^{k} p_c \log_2 p_c$. \textit{Higher entropy values} signify that CQs are more evenly distributed across the $k$ clusters, suggesting a \textit{greater diversity} of requirements within the set. 
\end{itemize}

\vspace{0.05in} 
\noindent \paragraph{Pairwise set comparisons}
focused on quantifying the semantic overlap between pairs of CQ sets (e.g., Set A vs. Set B). For each set pair, with $N_A$ and $N_B$ CQs respectively, we compute the following metrics.

\begin{itemize}[leftmargin=0pt,itemsep=2pt,topsep=0pt]
    \renewcommand\labelitemi{}
    \item \textbf{\textit{Centroid cosine similarity}} between the set centroids $\mathbf{\bar{e}}_A$ and $\mathbf{\bar{e}}_B$ provides a measure of the overall alignment of their central semantic themes. A score closer to 1 indicates that the two sets are, on average, focused on similar topics.

    \item \textbf{\textit{Coverage}} measures how well one set covers the content of another. This was performed in both directions. To illustrate, for the coverage of Set A by Set B:
    \begin{itemize}
        \item \textbf{Mean Maximum Similarity (MMS).} For each CQ embedding $\mathbf{e}_{A,i}$ in Set A, its maximum cosine similarity to any CQ embedding in Set B, $s_{A_i \rightarrow B} = \max_{j} \cos(\mathbf{e}_{A,i}, \mathbf{e}_{B,j})$, was identified. The mean of these $s_{A_i \rightarrow B}$ scores (and its standard deviation) indicates, on average, how well each CQ in Set A is semantically represented by its closest counterpart in Set B. A higher mean suggests stronger semantic parallels offered by Set B.
        \item \textbf{Set coverage and novelty.} The percentage of CQs in Set A for which $s_{A_i \rightarrow B} \geq \tau$ (where $\tau$ is a pre-defined similarity threshold) was calculated. This quantifies the proportion of Set A's semantic content considered adequately ``explained'' or represented by Set B.
        Consequently, the percentage novelty represents the proportion of Set A that introduces semantic content not found (or not closely matched via $\tau$) in Set B.
    \end{itemize}
    The same metrics were computed for the coverage of Set B by Set A.

    \item \textbf{\textit{Bidirectional coverage}} quantifies the overall mutual semantic overlap, given as: $\frac{N_{A \rightarrow B}^{\textnormal{cov}} + N_{B \rightarrow A}^{\textnormal{cov}}}{N_A + N_B}$, where $N_{A \rightarrow B}^{\textnormal{cov}}$ is the number of CQs in Set A covered by Set B (i.e., $s_{A_i \rightarrow B} \geq \tau$), and $N_{B \rightarrow A}^{\textnormal{cov}}$ is the number of CQs in Set B covered by Set A. A higher percentage indicates greater shared conceptual space between the sets.
\end{itemize}

The aim of this framework is to offer a practical, scalable approach to comparing CQs using established linguistic and semantic metrics.
The LLM-based relevance measure provides a proxy for expert judgment, reducing reliance on manual annotation.
Although readability metrics and general-purpose embeddings have limitations and depend on parameter choices, these affect comparably the results obtained from the datasets used in the complementary analyses.
While some subjectivity and model bias remain, the framework delivers consistent relative evaluations suitable for benchmarking across models and domains.

%% file: tables/readability_measures.tex
\begin{table}[t!]
\caption{Summary of readability measures used in this study. $|S|$, $|W|$, and $|Syl|$ denote the number of sentences, words, and syllables in the text, respectively. $|DC_{DW}|$ is the number of difficult words after excluding Dale-Chall's list of 3k common words.}
\label{tab:readabilityMeasures}
\begin{tabularx}{\textwidth}{|p{1.9cm}|p{3cm}|X|}
\hline
\textbf{Readability Measure}  & \textbf{Formula} & \textbf{Description} \\
\hline
Flesch-Kincaid Grade Level ({\bf FKGL}) \cite{kincaid1975} & $11.8 \times (\frac{|Syl|}{|W|}) + 0.39 \times (\frac{|W|}{|S|}) - 15.59$ & Estimates the years of formal education (U.S. grade, from kindergarten/nursery to college) a person needs to understand a piece of writing on the first reading. \\
\hline
Dale-Chall Readability Index ({\bf DCR}) \cite{dale1948} & $0.1579 \times (\frac{|DC_{DW}|}{|W|} \times 100) + 0.0496 \times (\frac{|W|}{|S|})$ & 
Estimates the comprehension difficulty of text based on word familiarity and sentence structure (using a specific list of 3,000 familiar words that are easily understood by 4th-grade students).
\\
\hline
\end{tabularx}
\end{table}

%% file: sections/results.tex
\section{Results}\label{sec:results}

This section presents the results of our comparative analysis of CQ datasets. We first report the outcomes of the analysis of CQ features (RQ1), followed by the set analysis of CQ embeddings (RQ2).
In all experiments, LLMs were prompted via their respective APIs using the following parameters to maximise reproducibility: $\text{temperature}=0$, $\text{Top-P}=1$, and $\text{seed}=46$.
All code and materials are available at (\url{https://github.com/KE-UniLiv/compcq}).

\subsection{Computational analysis of LLM-generated CQ feature (RQ1)}

\input{tables/features_table}

We compare the features of LLM-generated CQs (Section~\ref{sssec:cq-features}) across all requirements/domains.
The aggregated metrics are shown in Table~\ref{tab:aggregate_stats}.
%
The complexity of the source domain remains a primary determinant of CQ complexity, regardless of the model.
The \textit{Personalized Depression Treatment Ontology (PDTO)}, a highly complex, context-rich domain involving relationships between genetics, treatments, and outcomes, consistently elicited the most complex and least readable CQs from \textit{all five models}.
This domain produced the peak scores for nearly every model in every complexity and readability metric (e.g., $c_1$ of $16.33$ for \texttt{KimiK2}, FKGL of $18.21$ for \texttt{Llama3.2-3B}).
Conversely, the \textit{When to Go Where} domain, which describes a more straightforward recommendation task based on preferences, produced some of the simplest and most readable CQs.

\texttt{Gemini} consistently produced the most concise, simple, and readable CQs across the majority of domains.
For \textit{Music Meta}, \textit{BME}, and \textit{When to Go Where}, it registered the lowest scores in length ($c_0$), requirement complexity ($c_1$), lexical complexity ($c_2$), and syntactic complexity ($c_3$).
This low complexity translates directly to high readability, with \texttt{Gemini} achieving the best (lowest) FKGL scores in the same domains, suggesting a generation style that favours simple, direct questions.
In  contrast, \texttt{KimiK2} and \texttt{Llama3.2-3B} consistently generate the most complex, verbose, and least readable CQs.
\texttt{Llama3.2-3B} produced the longest CQs ($c_0$) in three of the five domains, peaking at 229.5 characters for \textit{Music Meta}.
KimiK2 demonstrated the highest \textit{requirement} complexity ($c_1$) in four of the five domains, including a peak score of $16.3$ for the \textit{PDTO} ontology.
This high complexity indicates that the CQs generated by these models demand a greater number of ontological primitives (concepts, properties, and relations) and are constructed with more elaborate phrasing and complex grammatical structures (e.g., high $c_2$ and $c_3$ scores).
Consequently, these models also produced the least readable CQs.
\texttt{Llama3.2-3B}, for instance, scored a high (poor) FKGL of 18.21 for the \textit{PDTO} domain, suggesting a reading level far beyond that of other models for the same task.
\texttt{GPT} and \texttt{Llama3.1-8B} occupy a middle ground.
GPT's output is generally more complex and less readable than Gemini's but significantly more streamlined than \texttt{KimiK2} or the \texttt{Llama 3.2} model.

Despite the variance in complexity, all models extracted highly \textit{relevant} CQs, with mean scores consistently above 3.
This indicates that their CQs are closely aligned with the source requirements, which suggests low hallucination. 
Proprietary models (\texttt{Gemini}, \texttt{GPT}) showed the most consistent performance, with highest relevance scores (often $> 3.9$) across all domains.
This suggests their outputs map almost exclusively the \textit{explicitly} stated requirements from the source material.
Open models showed minor, but notable, dips in specific domains (\texttt{KimiK2} scored $3$ in \textit{BME}, and \texttt{Llama3.2-3B} scored $3.4$ in \textit{When to Go Where}).

Overall, the results indicate that model output style is not fixed but varies systematically with the semantic 
complexity of the input domain. Open models, already biased toward verbose generations, exhibit disproportionate increases in output complexity when operating in more demanding domains.

\input{tables/iset_table}

\subsection{Set analysis: diversity and overlap (RQ2)}

We observed a significant disparity in the quantity of CQs generated (c.f. Table~\ref{tab:internal_diversity}).
\textit{Gemini}, \textit{GPT} and \textit{KimiK2} consistently produced a reasonable number of questions, typically between $15$ and $35$ CQs per ontology.
In contrast, \texttt{Llama} models generated a very low number of CQs ($8-15$).
This suggests a potential failure to capture the full breadth of the requirements.

\vspace{0.05in} 
\noindent\paragraph{Internal semantic diversity:}
The analysis of internal semantic diversity, presented in Table~\ref{tab:internal_diversity}, confirms the dependence on the specific ontology domain.
No single model was consistently the most or least diverse across all datasets.

Model performance demonstrated high variability.
\texttt{Llama3.2-3B}, for example, exhibited a ``boom-or-bust'' profile.
It produced the \textit{least} diverse set of CQs for the \textit{Music Meta} ontology, registering the highest average cosine similarity (0.65) and lowest centroid distance (0.56).
However, for \textit{When To Go Where} and \textit{Political Journalism}, it generated the \textit{most} diverse sets, with the lowest cosine similarities ($0.23$ and $0.275$, respectively) and highest centroid distances ($0.85$ and $0.79$).

\texttt{GPT} consistently produced CQs of \textit{low diversity}, suggesting its outputs are often tightly clustered around a few core topics.
It registered the least diverse set for the \textit{Political Journalism} ontology (avg. cosine sim $0.54$) and the second-least for \textit{Music Meta} (0.53).
Notably, its Shannon entropy scores were often high (e.g., 2.283 for \textit{PDTO}), indicating that while its CQs are semantically similar, they are at least evenly distributed across the $k=5$ sub-topics.
\texttt{Gemini} showed strong, high-diversity performance in some domains but was inconsistent.
It achieved the best diversity scores for \textit{BME} (ACS $0.23$, ACD $0.864$, Entropy $2.29$).
Conversely, it generated the \textit{least} diverse set of CQs for \textit{When To Go Where}  (ACS $0.509$).

Among the open LLMs, \texttt{KimiK2} emerged as a consistent model, often matching or exceeding the diversity of closed models (e.g., in \textit{BME} and \textit{PDTO}) while also generating a reasonable number of CQs.
Llama models, beyond their low output quantity, showed erratic entropy scores, suggesting they may fail to identify and cover all $k=5$ sub-topics (e.g., \texttt{Llama3.1-8B} in \textit{Music Meta}, wih entropy $1.82$).

In summary, the internal diversity of a generated CQ set is an emergent property of the specific model-domain interaction.
Models like \texttt{KimiK2} and \texttt{Llama3.2-3B} can achieve high diversity, but their performance is erratic.
GPT appears to be a low-diversity ``\textit{generalist}'', while \texttt{Gemini} and \texttt{KimiK2} show promise but are also subject to domain-specific variability.
The critically low CQ output from the Llama models, however, remains their most significant drawback -- casting doubt on their ability to comprehensively cover the requirements space.

\vspace{0.05in} 
\noindent\paragraph{Pairwise set comparison:}
The pairwise comparison of CQ sets confirms that semantic overlap is also dependent on the nature of the ontology domain.

We observed high convergence on constrained domains (Table~\ref{tab:pairwise_overlap_all}).
The \textit{When To Go Where} use case provides a stark example of convergence. This domain, which describes a clear, constrained recommendation task (finding parks by weather, crowd, and location), forced high thematic agreement among most models.
Centroid similarities were exceptionally high (e.g., \texttt{Gemini-GPT}:$0.91$, \texttt{Gemini-KimiK2}:0.88, \texttt{GPT-Llama3.1-8B}: $0.93$).
Notably, the bidirectional coverage was very high, indicating the models generated semantically equivalent CQs.
The overlap between \texttt{Gemini} and \texttt{GPT} was 56.2\%, and the overlap between \texttt{GPT} and \texttt{Llama3.1-8B} was 42.9\%.
This suggests that when a use case is narrow and well-defined, models tend to generate similar core requirements, resulting in low novelty.
The exception in this domain was \texttt{Llama3.2-3B}, which was a clear outlier.
It registered 0\% coverage with every other model and had exceptionally low centroid similarities (e.g., $0.25$ vs. \texttt{Gemini}).
This suggests it fundamentally misunderstood the primary requirements of the use case.

Conversely, high novelty was observed in broad domains.
The \textit{Music Meta} and \textit{BME} domains, where requirements are specified as personas and user stories, resulted in almost zero semantic overlap (0\% bidir. coverage between nearly all model pairs).
This indicates that while the models were thematically aligned (as shown by moderate-to-high centroid similarities), they each explored unique facets of the requirements.
Their CQs were highly novel, with almost no semantic duplication between sets.
This suggests that for broad or creative domains captured by user stories, no single model can capture the full requirements space.

\textit{PDTO} and \textit{Political Journalism} represent a middle ground. In these complex, technical domains, we again observe high thematic alignment (centroid sim. often $> 0.85$) but generally low semantic overlap.
However, a key pattern emerges: the most significant overlap occurs between the two closed-source models, \texttt{Gemini} and \texttt{GPT}.
They achieved 23.8\% bidirectional coverage for \textit{PDTO} and 17.6\% for \textit{Journalism}.
In contrast, most other pairings (especially those involving \texttt{KimiK2} or \texttt{Llama3.2-3B}) had 0\% overlap.
This suggests that closed models (\texttt{Gemini} and \texttt{GPT}) may share a more similar interpretation of the core requirements for these complex, non-creative tasks.
A notable exception is the 26.7\% overlap between \texttt{GPT} and \texttt{Llama3.1-8B} for the \textit{PDTO} domain, which indicates a strong and specific convergence between those two models for that particular task.

In summary, constrained domains force convergence, while broad domains promote high novelty.
Across all domains, the CQ diversity strongly suggests that relying on a single LLM, regardless of its individual diversity, is potentially insufficient to capture the full spectrum of requirements.

\input{tables/pset_table}

%% file: tables/features_table.tex
\begin{table}[t]
    \centering
    \setlength{\belowrulesep}{0pt}
    \caption{CQ feature comparison on the LLM-generated sets for each requirement.}
    \label{tab:aggregate_stats}
    \resizebox{\textwidth}{!}{%
    \begin{tabular}{lccccccc}
    \toprule
    Model & Length ($c_0$) & Req. ($c_1$) & Lex. ($c_2$) & Syn. ($c_3$) & FKGL & DCR & Rel. \\
    \midrule
    \rowcolor{gray!5} \multicolumn{8}{l}{\textbf{Music Meta}} \\
    Gemini & 85.5 $\pm$ 28.3 & 8.8 $\pm$ 2.8 & 11.4 $\pm$ 3.4 & 32.3 $\pm$ 10.6 & 8.0 $\pm$ 3.1 & 9.3 $\pm$ 2.0 & 3.9 $\pm$ 0.3 \\
    GPT & 124.1 $\pm$ 24.3 & 8.4 $\pm$ 2.1 & 16.1 $\pm$ 3.0 & 42.6 $\pm$ 7.7 & 12.0 $\pm$ 2.8 & 12.1 $\pm$ 1.8 & 4.0 $\pm$ 0.2 \\
    KimiK2 & 174.5 $\pm$ 36.8 & 11.6 $\pm$ 2.5 & 21.1 $\pm$ 5.6 & 60.1 $\pm$ 12.4 & 15.1 $\pm$ 3.2 & 12.1 $\pm$ 1.0 & 3.7 $\pm$ 0.5 \\
    L3.1-8B & 100.0 $\pm$ 33.5 & 6.9 $\pm$ 2.0 & 14.3 $\pm$ 4.0 & 34.5 $\pm$ 12.1 & 14.1 $\pm$ 3.3 & 12.8 $\pm$ 1.2 & 3.7 $\pm$ 0.5 \\
    L3.2-3B & 229.5 $\pm$ 37.3 & 8.5 $\pm$ 2.9 & 27.2 $\pm$ 5.5 & 71.0 $\pm$ 11.1 & 13.8 $\pm$ 3.8 & 12.5 $\pm$ 1.3 & 3.8 $\pm$ 0.4 \\
    \midrule

    \rowcolor{gray!5} \multicolumn{8}{l}{\textbf{BME}} \\
    Gemini & 87.0 $\pm$ 25.1 & 6.7 $\pm$ 2.8 & 10.3 $\pm$ 3.3 & 29.3 $\pm$ 8.9 & 9.6 $\pm$ 3.0 & 11.7 $\pm$ 1.6 & 3.9 $\pm$ 0.3 \\
    GPT & 103.2 $\pm$ 27.9 & 8.8 $\pm$ 3.1 & 13.2 $\pm$ 3.1 & 36.0 $\pm$ 11.1 & 11.5 $\pm$ 3.1 & 12.0 $\pm$ 1.9 & 3.9 $\pm$ 0.3 \\
    KimiK2 & 168.3 $\pm$ 39.6 & 11.2 $\pm$ 4.0 & 18.6 $\pm$ 4.2 & 52.2 $\pm$ 12.9 & 15.3 $\pm$ 2.5 & 13.1 $\pm$ 1.3 & 3.0 $\pm$ 1.0 \\
    L3.1-8B & 139.5 $\pm$ 19.3 & 7.8 $\pm$ 2.6 & 15.9 $\pm$ 1.4 & 41.4 $\pm$ 6.5 & 15.3 $\pm$ 2.1 & 13.0 $\pm$ 1.4 & 3.3 $\pm$ 1.0 \\
    L3.2-3B & 145.8 $\pm$ 23.3 & 9.4 $\pm$ 3.2 & 18.9 $\pm$ 2.8 & 49.2 $\pm$ 7.1 & 14.1 $\pm$ 2.6 & 10.9 $\pm$ 1.5 & 3.6 $\pm$ 0.8 \\
    \midrule

    \rowcolor{gray!5} \multicolumn{8}{l}{\textbf{When to Go Where}} \\
    Gemini & 89.1 $\pm$ 27.6 & 8.2 $\pm$ 3.0 & 12.1 $\pm$ 3.6 & 32.8 $\pm$ 9.5 & 8.1 $\pm$ 3.2 & 10.3 $\pm$ 1.7 & 4.0 $\pm$ 0.2 \\
    GPT & 112.2 $\pm$ 22.5 & 9.7 $\pm$ 3.1 & 14.2 $\pm$ 3.2 & 36.0 $\pm$ 8.6 & 10.8 $\pm$ 2.3 & 11.7 $\pm$ 1.6 & 4.0 $\pm$ 0.2 \\
    KimiK2 & 123.1 $\pm$ 43.3 & 9.2 $\pm$ 2.5 & 14.1 $\pm$ 4.6 & 42.6 $\pm$ 16.0 & 9.3 $\pm$ 3.2 & 12.2 $\pm$ 1.7 & 4.0 $\pm$ 0.0 \\
    L3.1-8B & 79.1 $\pm$ 21.3 & 8.3 $\pm$ 1.4 & 10.7 $\pm$ 3.7 & 26.5 $\pm$ 9.1 & 10.7 $\pm$ 2.0 & 10.6 $\pm$ 1.7 & 4.0 $\pm$ 0.0 \\
    L3.2-3B & 101.9 $\pm$ 20.1 & 6.1 $\pm$ 3.3 & 12.6 $\pm$ 1.6 & 32.9 $\pm$ 5.2 & 11.0 $\pm$ 2.6 & 12.0 $\pm$ 1.2 & 3.4 $\pm$ 1.0 \\
    \midrule

    \rowcolor{gray!5} \multicolumn{8}{l}{\textbf{Political Journalism Ontology}} \\
    Gemini & 104.1 $\pm$ 42.2 & 9.5 $\pm$ 3.3 & 12.6 $\pm$ 5.3 & 35.6 $\pm$ 14.8 & 10.5 $\pm$ 3.1 & 10.7 $\pm$ 1.6 & 3.9 $\pm$ 0.3 \\
    GPT & 103.9 $\pm$ 29.9 & 11.2 $\pm$ 3.1 & 12.6 $\pm$ 3.9 & 37.4 $\pm$ 10.9 & 13.3 $\pm$ 3.0 & 12.8 $\pm$ 1.8 & 3.9 $\pm$ 0.3 \\
    KimiK2 & 160.1 $\pm$ 29.6 & 13.4 $\pm$ 2.3 & 19.8 $\pm$ 4.0 & 55.2 $\pm$ 12.0 & 14.2 $\pm$ 2.9 & 12.8 $\pm$ 1.1 & 3.8 $\pm$ 0.4 \\
    L3.1-8B & 124.9 $\pm$ 30.9 & 8.9 $\pm$ 3.6 & 15.5 $\pm$ 3.4 & 40.0 $\pm$ 8.9 & 15.2 $\pm$ 2.9 & 11.7 $\pm$ 0.9 & 3.6 $\pm$ 1.0 \\
    L3.2-3B & 104.2 $\pm$ 68.7 & 9.6 $\pm$ 9.4 & 14.0 $\pm$ 9.1 & 37.2 $\pm$ 25.0 & 11.9 $\pm$ 4.7 & 12.2 $\pm$ 1.2 & 3.7 $\pm$ 1.0 \\
    \midrule

    \rowcolor{gray!5} \multicolumn{8}{l}{\textbf{Personalized Depression Treatment Ontology}} \\
    Gemini & 137.8 $\pm$ 32.3 & 12.5 $\pm$ 3.6 & 16.5 $\pm$ 4.3 & 44.6 $\pm$ 12.6 & 13.9 $\pm$ 3.0 & 12.8 $\pm$ 1.4 & 3.9 $\pm$ 0.3 \\
    GPT & 119.7 $\pm$ 19.7 & 9.8 $\pm$ 3.3 & 13.6 $\pm$ 2.6 & 35.9 $\pm$ 6.6 & 13.7 $\pm$ 2.2 & 12.8 $\pm$ 1.2 & 3.9 $\pm$ 0.3 \\
    KimiK2 & 171.8 $\pm$ 25.6 & 16.3 $\pm$ 5.5 & 19.6 $\pm$ 2.4 & 56.8 $\pm$ 9.9 & 16.2 $\pm$ 2.2 & 14.1 $\pm$ 1.5 & 3.6 $\pm$ 0.6 \\
    L3.1-8B & 115.6 $\pm$ 21.9 & 10.1 $\pm$ 3.4 & 13.7 $\pm$ 3.3 & 35.7 $\pm$ 9.1 & 12.7 $\pm$ 1.9 & 13.0 $\pm$ 1.3 & 3.6 $\pm$ 0.7 \\
    L3.2-3B & 176.6 $\pm$ 19.7 & 9.8 $\pm$ 4.1 & 19.7 $\pm$ 1.8 & 50.4 $\pm$ 5.2 & 18.2 $\pm$ 1.9 & 13.8 $\pm$ 1.0 & 3.7 $\pm$ 0.5 \\
    \bottomrule
    \end{tabular}%
    }
\end{table}

%% file: tables/iset_table.tex
\begin{table}[t]
    \centering
    \caption{Analysis of internal semantic diversity for CQ sets generated by each model across the five datasets. Lower `Avg. Cosine Sim' (ACS) and higher `Avg. Dist. to Centroid' (ACD) and `Shannon Entropy' ($H$) indicate greater diversity.}
    \label{tab:internal_diversity}

    \resizebox{\linewidth}{!}{%
    
    \begin{tabular}{lcccc @{\quad} lcccc}
    
    \toprule
    Model & CQs & ACS & ACD & H & 
    Model & CQs & ACS & ACD & H \\
    \midrule

    \multicolumn{5}{l}{\textbf{Music Meta}} & \multicolumn{5}{l}{\textbf{BME}} \\
    
    Gemini      & 31 & 0.45 $\pm$ 0.18 & 0.72 $\pm$ 0.14 & 2.06 & Gemini      & 35 & 0.23 $\pm$ 0.17 & 0.86 $\pm$ 0.06 & 2.29 \\
    GPT         & 20 & 0.53 $\pm$ 0.14 & 0.66 $\pm$ 0.11 & 2.20 & GPT         & 25 & 0.31 $\pm$ 0.14 & 0.81 $\pm$ 0.06 & 2.23 \\
    KimiK2      & 15 & 0.36 $\pm$ 0.09 & 0.77 $\pm$ 0.06 & 2.01 & KimiK2      & 24 & 0.23 $\pm$ 0.12 & 0.86 $\pm$ 0.05 & 2.20 \\
    L3.1-8B     & 13 & 0.40 $\pm$ 0.20 & 0.74 $\pm$ 0.09 & 1.82 & L3.1-8B     & 13 & 0.37 $\pm$ 0.11 & 0.76 $\pm$ 0.05 & 2.20 \\
    L3.2-3B     & 11 & 0.65 $\pm$ 0.12 & 0.56 $\pm$ 0.10 & 1.87 & L3.2-3B     & 15 & 0.42 $\pm$ 0.12 & 0.74 $\pm$ 0.04 & 2.18 \\
    \midrule

    \multicolumn{5}{l}{\textbf{When To Go Where}} & \multicolumn{5}{l}{\textbf{Political Journalism}} \\

    Gemini      & 28 & 0.51 $\pm$ 0.13 & 0.68 $\pm$ 0.08 & 2.24 & Gemini      & 24 & 0.42 $\pm$ 0.15 & 0.74 $\pm$ 0.09 & 1.95 \\
    GPT         & 20 & 0.44 $\pm$ 0.23 & 0.71 $\pm$ 0.15 & 1.98 & GPT         & 27 & 0.54 $\pm$ 0.15 & 0.66 $\pm$ 0.10 & 2.22 \\
    KimiK2      & 17 & 0.43 $\pm$ 0.15 & 0.73 $\pm$ 0.07 & 1.97 & KimiK2      & 20 & 0.36 $\pm$ 0.13 & 0.78 $\pm$ 0.07 & 2.15 \\
    L3.1-8B     & 8  & 0.32 $\pm$ 0.19 & 0.76 $\pm$ 0.10 & 2.16 & L3.1-8B     & 9  & 0.42 $\pm$ 0.22 & 0.71 $\pm$ 0.09 & 2.06 \\
    L3.2-3B     & 16 & 0.23 $\pm$ 0.14 & 0.85 $\pm$ 0.04 & 2.10 & L3.2-3B     & 9  & 0.28 $\pm$ 0.22 & 0.79 $\pm$ 0.12 & 2.06 \\
    \midrule

    \multicolumn{5}{l}{\textbf{Personalized Depression Treatment}} & \multicolumn{5}{l}{} \\ 

    Gemini      & 20 & 0.41 $\pm$ 0.11 & 0.74 $\pm$ 0.06 & 2.08 & & & & & \\ 
    GPT         & 22 & 0.46 $\pm$ 0.12 & 0.72 $\pm$ 0.05 & 2.28 & & & & & \\
    KimiK2      & 15 & 0.34 $\pm$ 0.11 & 0.79 $\pm$ 0.07 & 1.84 & & & & & \\
    L3.1-8B     & 8  & 0.42 $\pm$ 0.14 & 0.71 $\pm$ 0.05 & 2.00 & & & & & \\
    L3.2-3B     & 10 & 0.42 $\pm$ 0.12 & 0.72 $\pm$ 0.08 & 2.05 & & & & & \\
    
    \bottomrule
    \end{tabular}%
    }
\end{table}

%% file: tables/pset_table.tex
{ 
\newcommand{\stdev}[1]{$\pm$#1}
\footnotesize
\setlength{\tabcolsep}{1pt}
\setlength{\belowrulesep}{0pt}
\begin{xltabular}{\linewidth}{X c cc cc c} 
    
    \caption{Pairwise semantic overlap across LLM-generated CQ sets.}
    \label{tab:pairwise_overlap_all} \\

    \toprule
    \textbf{Sets} & \textbf{Centroid} & \multicolumn{2}{c}{\textbf{S1 $\leftarrow$ S2}} & \multicolumn{2}{c}{\textbf{S1 $\rightarrow$ S2}} & \textbf{BiDirect.} \\
    \textbf{(S1 $\leftrightarrow$ S2)} & \textbf{Sim.} & \textbf{Cov.(\%)} & \textbf{MMS} & \textbf{Cov.(\%)} & \textbf{MMS} & \textbf{Cov.(\%)} \\
    \midrule
    \endfirsthead 
    
    \multicolumn{7}{c}%
    {{\tablename\ \thetable{} -- continued from previous page}} \\
    \toprule
    \textbf{Sets} & \textbf{Centroid} & \multicolumn{2}{c}{\textbf{S1 $\leftarrow$ S2}} & \multicolumn{2}{c}{\textbf{S1 $\rightarrow$ S2}} & \textbf{BiDirect.} \\
    \textbf{(S1 $\leftrightarrow$ S2)} & \textbf{Sim.} & \textbf{Cov.(\%)} & \textbf{MMS} & \textbf{Cov.(\%)} & \textbf{MMS} & \textbf{Cov.(\%)} \\
    \midrule
    \endhead 
    
    \bottomrule
    \endlastfoot 

    
    \rowcolor{gray!5} \multicolumn{7}{l}{\textbf{Music Meta}} \\
    Gemini, GPT & 0.57 &0& 0.47 \stdev{0.11} &0& 0.54 \stdev{0.06} &0\\
    Gemini, KimiK2 & 0.57 &0& 0.43 \stdev{0.11} &0& 0.49 \stdev{0.09} &0\\
    Gemini, L3.1-8B & 0.37 &0& 0.35 \stdev{0.12} &0& 0.38 \stdev{0.17} &0\\
    Gemini, L3.2-3B & 0.37 &0& 0.28 \stdev{0.12} &0& 0.49 \stdev{0.04} &0\\
    GPT, KimiK2 & 0.84 &0& 0.61 \stdev{0.06} &0& 0.57 \stdev{0.08} &0\\
    GPT, L3.1-8B & 0.79 & 30.0 & 0.67 \stdev{0.10} & 38.5 & 0.60 \stdev{0.17} & 33.3 \\
    GPT, L3.2-3B & 0.77 & 10.0 & 0.57 \stdev{0.14} & 27.3 & 0.70 \stdev{0.07} & 16.1 \\
    KimiK2, L3.1-8B & 0.69 &0& 0.49 \stdev{0.10} &0& 0.49 \stdev{0.12} &0\\
    KimiK2, L3.2-3B & 0.71 &0& 0.47 \stdev{0.11} &0& 0.58 \stdev{0.05} &0\\
    L3.1-8B, L3.2-3B & 0.74 &0& 0.50 \stdev{0.17} &0& 0.64 \stdev{0.05} &0\\
    \midrule

    \rowcolor{gray!5} \multicolumn{7}{l}{\textbf{BME}} \\
    Gemini, GPT & 0.74 &0& 0.48 \stdev{0.12} &0& 0.51 \stdev{0.11} &0\\
    Gemini, KimiK2 & 0.54 &0& 0.35 \stdev{0.11} &0& 0.40 \stdev{0.09} &0\\
    Gemini, L3.1-8B & 0.39 & 2.9 & 0.32 \stdev{0.15} & 7.7 & 0.47 \stdev{0.13} & 4.2 \\
    Gemini, L3.2-3B & 0.56 &0& 0.39 \stdev{0.13} &0& 0.46 \stdev{0.09} &0\\
    GPT, KimiK2 & 0.70 &0& 0.48 \stdev{0.09} &0& 0.45 \stdev{0.11} &0\\
    GPT, L3.1-8B & 0.65 &0& 0.53 \stdev{0.09} &0& 0.57 \stdev{0.09} &0\\
    GPT, L3.2-3B & 0.80 & 4.0 & 0.54 \stdev{0.11} & 6.7 & 0.56 \stdev{0.09} & 5.0 \\
    KimiK2, L3.1-8B & 0.74 &0& 0.44 \stdev{0.10} &0& 0.51 \stdev{0.10} &0\\
    KimiK2, L3.2-3B & 0.69 &0& 0.42 \stdev{0.09} &0& 0.47 \stdev{0.09} &0\\
    L3.1-8B, L3.2-3B & 0.71 & 7.7 & 0.56 \stdev{0.13} & 6.7 & 0.53 \stdev{0.13} & 7.1 \\
    \midrule

    \rowcolor{gray!5} \multicolumn{7}{l}{\textbf{When To Go Where}} \\
    Gemini, GPT & 0.91 & 57.1 & 0.77 \stdev{0.09} & 55.0 & 0.69 \stdev{0.21} & 56.2 \\
    Gemini, KimiK2 & 0.88 & 28.6 & 0.72 \stdev{0.08} & 35.3 & 0.72 \stdev{0.13} & 31.1 \\
    Gemini, L3.1-8B & 0.85 & 17.9 & 0.67 \stdev{0.10} & 50.0 & 0.66 \stdev{0.22} & 25.0 \\
    Gemini, L3.2-3B & 0.25 &0& 0.42 \stdev{0.08} &0& 0.23 \stdev{0.18} &0\\
    GPT, KimiK2 & 0.87 & 30.0 & 0.65 \stdev{0.12} & 35.3 & 0.69 \stdev{0.09} & 32.4 \\
    GPT, L3.1-8B & 0.93 & 35.0 & 0.68 \stdev{0.14} & 62.5 & 0.77 \stdev{0.07} & 42.9 \\
    GPT, L3.2-3B & 0.47 &0& 0.49 \stdev{0.08} &0& 0.41 \stdev{0.17} &0\\
    KimiK2, L3.1-8B & 0.82 &0& 0.60 \stdev{0.06} &0& 0.61 \stdev{0.07} &0\\
    KimiK2, L3.2-3B & 0.29 &0& 0.42 \stdev{0.10} &0& 0.27 \stdev{0.17} &0\\
    L3.1-8B, L3.2-3B & 0.49 &0& 0.54 \stdev{0.10} &0& 0.38 \stdev{0.16} &0\\
    \midrule

    \rowcolor{gray!5} \multicolumn{7}{l}{\textbf{Political Journalism Ontology}} \\
    Gemini, GPT & 0.87 & 16.7 & 0.67 \stdev{0.09} & 18.5 & 0.69 \stdev{0.07} & 17.6 \\
    Gemini, KimiK2 & 0.91 & 4.2 & 0.63 \stdev{0.09} & 5.0 & 0.58 \stdev{0.12} & 4.5 \\
    Gemini, L3.1-8B & 0.67 &0& 0.50 \stdev{0.09} &0& 0.56 \stdev{0.10} &0\\
    Gemini, L3.2-3B & 0.53 &0& 0.43 \stdev{0.12} &0& 0.47 \stdev{0.16} &0\\
    GPT, KimiK2 & 0.86 & 3.7 & 0.60 \stdev{0.08} & 5.0 & 0.57 \stdev{0.08} & 4.3 \\
    GPT, L3.1-8B & 0.76 &0& 0.59 \stdev{0.10} &0& 0.60 \stdev{0.14} &0\\
    GPT, L3.2-3B & 0.64 &0& 0.51 \stdev{0.08} &0& 0.49 \stdev{0.17} &0\\
    KimiK2, L3.1-8B & 0.68 &0& 0.44 \stdev{0.12} &0& 0.55 \stdev{0.09} &0\\
    KimiK2, L3.2-3B & 0.57 &0& 0.41 \stdev{0.09} &0& 0.41 \stdev{0.09} &0\\
    L3.1-8B, L3.2-3B & 0.70 &0& 0.51 \stdev{0.09} &0& 0.49 \stdev{0.10} &0\\
    \midrule

    \rowcolor{gray!5} \multicolumn{7}{l}{\textbf{Personalized Depression Treatment Ontology}} \\
    Gemini, GPT & 0.89 & 25.0 & 0.66 \stdev{0.11} & 22.7 & 0.65 \stdev{0.11} & 23.8 \\
    Gemini, KimiK2 & 0.85 &0& 0.55 \stdev{0.08} &0& 0.52 \stdev{0.11} &0\\
    Gemini, L3.1-8B & 0.85 & 5.0 & 0.61 \stdev{0.08} & 12.5 & 0.61 \stdev{0.11} & 7.1 \\
    Gemini, L3.2-3B & 0.53 &0& 0.41 \stdev{0.10} &0& 0.42 \stdev{0.12} &0\\
    GPT, KimiK2 & 0.76 &0& 0.50 \stdev{0.10} &0& 0.49 \stdev{0.11} &0\\
    GPT, L3.1-8B & 0.87 & 18.2 & 0.65 \stdev{0.11} & 50.0 & 0.74 \stdev{0.10} & 26.7 \\
    GPT, L3.2-3B & 0.67 &0& 0.50 \stdev{0.07} &0& 0.55 \stdev{0.12} &0\\
    KimiK2, L3.1-8B & 0.70 &0& 0.43 \stdev{0.13} &0& 0.49 \stdev{0.09} &0\\
    KimiK2, L3.2-3B & 0.42 &0& 0.32 \stdev{0.07} &0& 0.29 \stdev{0.09} &0\\
    L3.1-8B, L3.2-3B & 0.65 & 12.5 & 0.53 \stdev{0.14} & 10.0 & 0.51 \stdev{0.16} & 11.1 \\

\end{xltabular} 
} 

%% file: sections/discussion+conclusion.tex
\section{Conclusions}\label{sec:conclusions}

LLMs are increasingly proposed as a solution to the knowledge acquisition bottleneck, to automatise laborious ontology engineering tasks, such as CQ generation.
However the intrinsic characteristics of the resulting CQs has not been fully investigated.
To address this gap, we introduce a new framework (\textsc{CompCQ}) that supports cross-domain empirical studies and systematically quantifies the readability, complexity, relevance, and semantic diversity of different sets of CQs.

Our cross-domain analysis of generated CQ confirms that the generation behaviour is a complex interplay between the model's intrinsic profile and the specific nature of the requirements.
Our experiments show that domain characteristics, and therefore the way requirements are captured and represented influence both the complexity of the output and the degree of semantic agreement between models.
In addition, each model exhibits a \textbf{distinct and measurable ``generation profile''} with clear trade-offs, highlighting a divide between closed and open models, which is potentially explained by their increased capacity (no. of parameters) and training material.
The high novelty across the board suggests that ensemble-based approaches could potentially increase requirement diversity.
Ontology engineers should leverage multiple models (e.g., \texttt{Gemini} for simple, core CQs; \texttt{KimiK2} for diverse, exploratory CQs) to brainstorm a comprehensive initial set.
This automated generation must then be followed by crucial human-in-the-loop refinement to curate, clarify, and augment this set, ensuring all functional requirements are fully captured.